\documentclass[10pt,journal,compsoc]{IEEEtran}
\usepackage{scalerel}
\usepackage{tikz}
\usetikzlibrary{svg.path}

\usepackage{graphicx}
\usepackage{amsfonts}
\usepackage{mathrsfs}
\usepackage{amsmath} 

\usepackage{scalerel}

\usepackage{bm}
\usepackage{booktabs,ragged2e}

\usepackage[flushleft]{threeparttable}

\usepackage{tabularx}
\newcolumntype{C}{>{\centering\arraybackslash}X} 
\usepackage{threeparttable}
\usepackage{multirow}
\usepackage[utf8]{inputenc}
\usepackage{newunicodechar}
\usepackage{booktabs}

\usepackage{scalerel}
\usepackage{tikz}
\usetikzlibrary{svg.path}

\definecolor{orcidlogocol}{HTML}{A6CE39}
\tikzset{
  orcidlogo/.pic={
    \fill[orcidlogocol] svg{M256,128c0,70.7-57.3,128-128,128C57.3,256,0,198.7,0,128C0,57.3,57.3,0,128,0C198.7,0,256,57.3,256,128z};
    \fill[white] svg{M86.3,186.2H70.9V79.1h15.4v48.4V186.2z}
                 svg{M108.9,79.1h41.6c39.6,0,57,28.3,57,53.6c0,27.5-21.5,53.6-56.8,53.6h-41.8V79.1z M124.3,172.4h24.5c34.9,0,42.9-26.5,42.9-39.7c0-21.5-13.7-39.7-43.7-39.7h-23.7V172.4z}
                 svg{M88.7,56.8c0,5.5-4.5,10.1-10.1,10.1c-5.6,0-10.1-4.6-10.1-10.1c0-5.6,4.5-10.1,10.1-10.1C84.2,46.7,88.7,51.3,88.7,56.8z};
  }
}

\newcommand\orcidicon[1]{\href{https://orcid.org/#1}{\mbox{\scalerel*{
\begin{tikzpicture}[yscale=-1,transform shape]
\pic{orcidlogo};
\end{tikzpicture}
}{|}}}}

\usepackage{hyperref} 

\usepackage{graphicx}
\usepackage{lipsum}

\usepackage{scalerel}
\usepackage[ruled,vlined]{algorithm2e}

\ifCLASSOPTIONcompsoc
  \usepackage[nocompress]{cite}
\else
  \usepackage{cite}
\fi
\ifCLASSINFOpdf
\else
\fi
\begin{document}
%
\title{Deep Transfer Learning with Graph Neural Network for Sensor-Based Human Activity Recognition}
%
%
%
%

\author{Yan~Yan$^\dag$$^{\orcidicon{0000-0002-6344-136X}}$, ~\IEEEmembership{Member,~IEEE,} Tianzheng~Liao$^\dag$, Jinjin~Zhao, Jiahong~Wang, Liang~Ma,  Wei~Lv, Jing~Xiong$^*$, and~Lei~Wang$^*$$^{\orcidicon{0000-0002-7033-9806}}$,~\IEEEmembership{Senior Member,~IEEE}
\IEEEcompsocitemizethanks{\IEEEcompsocthanksitem Y. Yan, T. Liao, J. Zhao, L. Ma, J. Xiong and L. Wang are from the Shenzhen Institutes of Advanced Technology, Chinese Academy of Sciences. T. Liao, W. Lv are from the City University of Macau. J. Zhao is also from the Northeastern University. J. Wang is from the General Hospital of People’s Liberation Army. (E-mail: yan.yan@siat.ac.cn)\protect\\
}
\thanks{$^\dag$ indicates joint first author, $^*$ indicates the corresponding authors.}
\thanks{ }}

\markboth{~}%
{Shell \MakeLowercase{\textit{et al.}}: Bare Demo of IEEEtran.cls for Computer Society Journals}
%



\IEEEtitleabstractindextext{%
\begin{abstract}
The sensor-based human activity recognition (HAR) in mobile application scenarios is often confronted with sensor modalities variation and annotated data deficiency. Given this observation, we devised a graph-inspired deep learning approach toward the sensor-based HAR tasks, which was further used to build a deep transfer learning model toward giving a tentative solution for these two challenging problems. Specifically, we present a multi-layer residual structure involved graph convolutional neural network (ResGCNN) toward the sensor-based HAR tasks, namely the HAR-ResGCNN approach. Experimental results on the PAMAP2 and mHealth data sets demonstrate that our ResGCNN is effective at capturing the characteristics of actions with comparable results compared to other sensor-based HAR models (with an average accuracy of 98.18\% and 99.07\%, respectively). More importantly, the deep transfer learning experiments using the ResGCNN model show excellent transferability and few-shot learning performance. The graph-based framework shows good meta-learning ability and is supposed to be a promising solution in sensor-based HAR tasks.
\end{abstract}

}

\maketitle

\IEEEdisplaynontitleabstractindextext

%
\IEEEpeerreviewmaketitle

\IEEEraisesectionheading{\section{Introduction}\label{sec:introduction}}
\IEEEPARstart{H}{uman} activity recognition (HAR) plays a vital role in human-machine interactions, mobile computing, and biomedical \& healthcare applications, which enables machines the ability to track the human activity state. 
HAR systems tracks the activity states through processing and learning information from some carriers that can record human actions (such as cameras \cite{liu2019tpami}, sensors\cite{ye2020harpfta}, radars\cite{ding2021radar}, WiFi signals\cite{huang2021wifi}, etc.). 
These activities include daily behaviors activities such as walking, running, lying, going upstairs, falling, sitting still, standing, and involve more complex activity classes such as roping, cycling, and jogging in sports scenarios.
In recent years, the rapid development and popularization of portable wearable sensors made the application scenarios of HAR more life-like and routine. 
The signal obtained by the wearable sensor can be used to analyze the activity status of the subject and determine the current user status.
Further, the activity information can also be well combined with physiological information (such as heart rate, blood oxygen sensor) and applied to the elderly care, remote health diagnosis, and daily fitness plan formulation \cite{andreu2015tbme}.

HAR using machine vision is a prevailing direction, which captures images or video streams to detect the behavior of humans with image/video processing technologies, such as \cite{jalal2014depth, ji20123d, jalal2017robust, simonyan2015two}, which have achieved good results in the field of video-based HAR. However, this method is restricted with the impact brought with the complex scenarios, the uncertainty of the action, and needs to consider the privacy problems caused by the camera, and is only suitable for some specific scenes. 
In contrast, wearable sensors are less susceptible to environmental interference, and the collected signals are more continuous and accurate and can be used in a broader range of scenarios. 
Over the past decade, sensor technology has made extraordinary advances in several areas, including computing power, size, precision, and manufacturing costs. 
These advances have enabled most of the sensors to be integrated into smartphones and other portable devices, making these devices more intelligent and more practical \cite{strac2021npj}.  
Wearable sensors commonly used for HAR are accelerometers, magnetometers, gyroscopes, and integrated inertial measurement units (IMUs). Typical works are \cite{shoaib2016complex, guo2016smartphone}.

For a long time, as a typical pattern recognition problem, many traditional machine learning algorithms have been used to solve the problems in sensor-based HAR, including decision tree \cite{fan2013human}, random forest \cite{hu2018novel}, support vector machine \cite{chathuramali2012faster}, Bayesian network \cite{xiao2018action}, Markov model \cite{sok2018activity} and so on. 
Under the strict control environment and limited input, the traditional maximum appearance algorithm has obtained good classification results. 
By extracting the time domain characteristics of simple statistical features and coefficients of time series analysis, SVM is used to classify 15 activities \cite{khan2014activity}, and a spinal cord injury patient HAR framework based on HMM is introduced in \cite{sok2018activity}.

However, traditional approaches with handcraft features were time-consuming, and the extracted features lack incremental \& unsupervised learning ability and generalization ability \cite{munoz2019outlier}, so the research based on deep learning has gradually achieved excellent results in this field and occupied a dominant position. 
The pre-processing of features is significantly reduced by the automatic extraction of features through multi-layer neural networks, while the deep learning structure has been proved to work well in unsupervised learning and intensive learning \cite{wang2019deep}. 
The automatic feature learning and classification using 5-layer hidden DNN has achieved good results \cite{hammerla2016deep}. 
\cite{ronao2015deep, zebin2016human, yang2015deep, ignatov2018real} confirmed that a convolutional neural network could be well used in the field of human motion recognition.
HAR models developed using LSTM\cite{guan2017lstm} and Bi-LSTM\cite{torres2018} were implemented with mobile phone sensors to make a good classification of daily life movements.

Although the deep learning techniques based on CNN, LSTM mentioned above have continued to develop in biomedical computing fields, the main adaption range of their application is limited to traditional Euclidean space.
The translational invariance assumptions restrict the abilities to express themselves in non-Euclidean spatial data analysis tasks \cite{wu2020gnnreview}. 
Recently, graph-based models were proposed to solve such spatial data analysis problems in non-Euclidean spaces, bringing a novel direction for deep learning.
Scarselli et al.\cite{scarselli2009gnn} proposed a graph neural network (GNN) model capable of handling multiple types of graphs, which designed a function that mapped the graph and its nodes to higher-dimensional Euclidean space and proposed an updated supervisory learning algorithm to estimate the model parameters of GNN and make a pioneering contribution to the subsequent development of GNN. 
Later, Bruna et al. \cite{Henaff2015DeepCN} combined the convolution ideas in spectrum theory with GNN and proposed graph convolution neural networks (GCNN). At present, the graph convolution network has been applied to many fields. Yu et al. \cite{yu2018stgcn} proposed a spatial time graph convolutional neural network (STGCN) deep learning approach to solve the problem of timing prediction in the field of transportation. Kearnes et al. \cite{kearnes2016} using the GCNN to encode atoms, keys, and distances can make better use of the information in the graph structure to provide a new paradigm of ligand-based virtual screening. Leskovec et al. \cite{ying2018} proposes a data-efficient GCNN algorithm for recommendation system Pinsage, which produces embedded expression on commodity nodes.
Since the cooperated body parts form the human activity processes, the correlation of different sensors in each allocated position revealed the variations when performing different activities.
Thus, the correlation-based graph structure information extracted from the sensing signals from the human body could be used in HAR tasks.
The multi-dimensional sensor signals might be investigated from the graph state transition viewpoint with correlations calculation mapping.
Considering that the traditional deep learning models cannot track such correlations, we propose a sensor-based HAR framework developed from graph neural network (GNN) to investigate human physical activities, which could learn the potential relationship with sensor-based information.

On the other hand, traditional machine learning models were often designed to solve specific tasks, requiring re-training when the task dataset distribution changes. 
The transfer learning idea is proposed to move beyond the isolated learning paradigm into using knowledge gained from one task to solve other related and similar tasks. 
In previous studies, the transfer learning paradigm were often used for natural language processing \cite{al2006}, medicine \cite{mua2019sensors}, bioinformatics, transportation science and recommendation system .
In the HAR tasks, especially in the real-world mobile health care or state monitoring applications, there are two notable challenges: firstly, the collected datasets usually have quite different data distributions, which restricts the adaptation of the non-transfer models; secondly, the lacking of labeled data resource restricts the constructing of practical models.
We propose a deep transfer learning model built with GNN to tackle the problems. The significances of this work are:
\begin{enumerate}
\item We propose a sensor-based HAR approach HAR-ResGCNN developed from GNN and taken advantage of the residual structures and Chebyshev graph filtering functions, which we termed as a ResGCNN framework. The HAR-ResGCNN approach accomplished excellent classification ability with overall accuracies of 98.18\% and 99.07\% performed on two public benchmarks sensor-based HAR datasets of PAMAP2 and mHealth, respectively.
\item With the ResGCNN framework, we perform the deep transfer learning experiments with cross-dataset settings. We illustrate that the transfer learning of ResGCNN shows excellent transferability with a faster convergence speed and a better learning curve than the non-transfer setting.
\item We further validate the ResGCNN-based transfer learning in few-shot cases. With limited labeled activity samples, the transferred ResGCNN shows good meta-learning ability and recognition performance, which proved to be an ideal model for dealing with applications having a small labeled set as in mobile computing.
\end{enumerate}

The rest of this work presents as follows: Section 2 describes the preliminaries of GNN and transfer learning; Section 3 illustrates the structure of HAR-ResGCNN and technique details; Section 4 proposes our materials and experiments; Section 5 presents and discusses the experimental results. Finally, Section 6 concludes this work.

\begin{figure}[]
\centering
\includegraphics[width=3.2in]{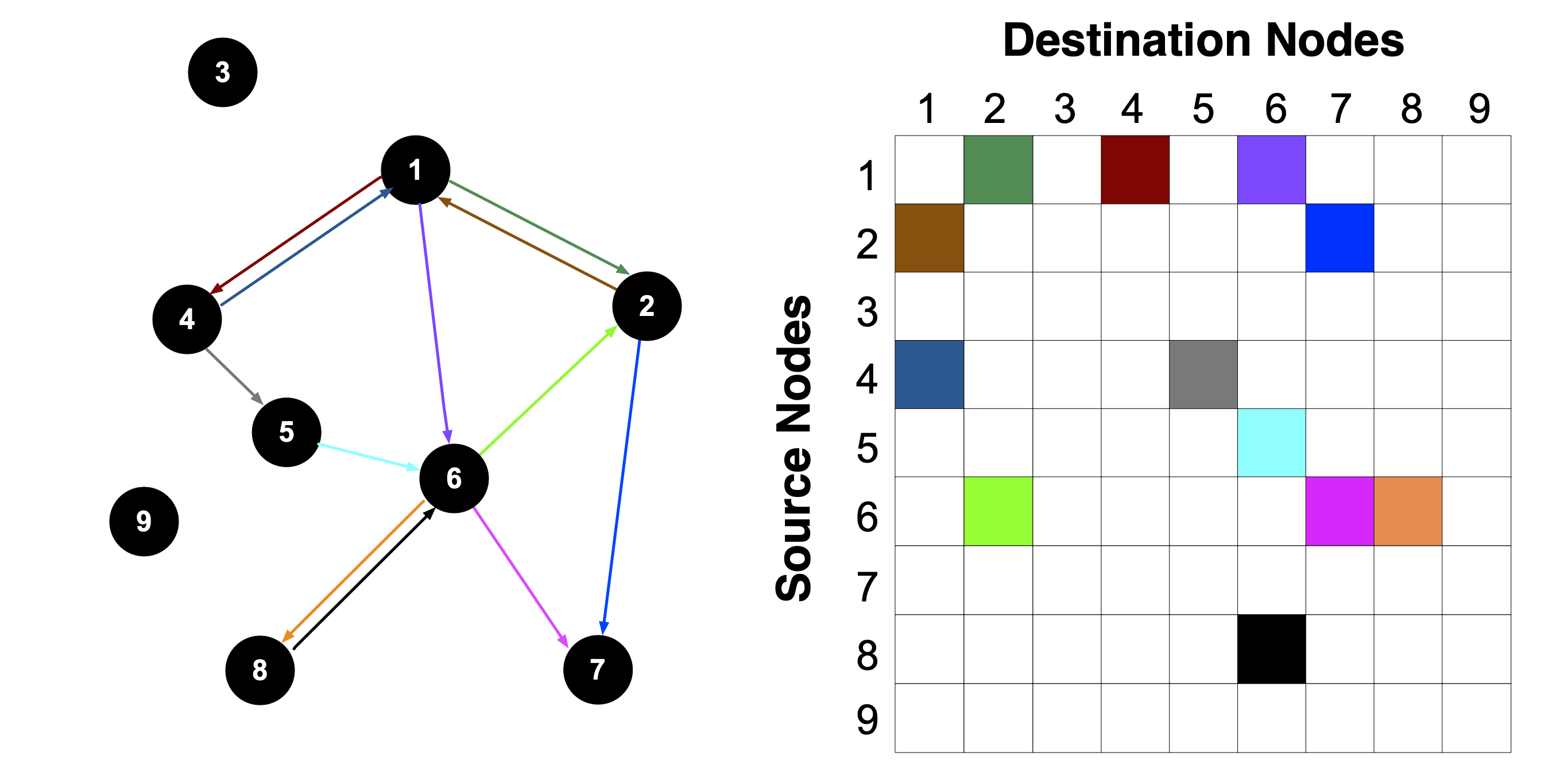}
\caption{This figure show changed into an undirected graph.}
\label{fig_graph}
\end{figure}

\section{Preliminaries}\label{sec:preliminaries}
This section introduces the preliminary knowledge of graph representation with spectral graph filtering and some related works using graph neural network learning models.
And then, we give the transfer learning approaches knowledge and related works.

\subsection{Graph Neural Network Preliminaries}
\subsubsection{Graph Representation}

A graph illustrated with $\mathcal{G} = \{\mathcal{V}, \mathcal{\xi}\}$ has $N$ vertices as a set of $\mathcal{V} = \{v_1, v_2, \cdots, v_N\}$ and $M$ edges as a set of $\mathcal{\xi} = \{e_1, e_2, \cdots, e_M\}$. 
The scale of the graphs is denoted with the number of vertices of $|\mathcal{V}|$, while the edges in $\mathcal{\xi}$ denote their connection relationship.
A graph might be undirected if an edge $e_i$ has two vertices of $v_{e_i}^1, v_{e_i}^2$ satisfies the equation of $e_i = (v_{e_i}^1, v_{e_i}^2) = (v_{e_i}^2, v_{e_i}^1)$, while directed when $(v_{e_i}^1, v_{e_i}^2) \neq (v_{e_i}^2, v_{e_i}^1)$.
Given a graph $\mathcal{G} = \{\mathcal{V}, \mathcal{\xi}\}$, it can be equivalently denoted with an adjacent matrix $\bm{A} \in \{0, 1\}^{N \times N}$.
The elements of $\bm{A}_{i, j}$ denotes the connection between $v_i$ and $v_j$, namely $\bm{A}_{i, j} = 1$ when adjacent, and $\bm{A}_{i, j} = 0$ when distant. 
A concrete example is illustrated in Figure \ref{fig_graph}, the value of the matrix is decided by the direction and weight of the edge between two different nodes.
A non-directed graph ignores the directions and turns a symmetric matrix, which is the considered type of graph data in this work.

\subsubsection{Graph Neural Networks}
GNNs use deep neural networks in the graph structural data analysis tasks.
The early GNN proposed in \cite{scarselli2005gnn, scarselli2008gnnntnnls} were used to learn both the features of nodes and graph structure. 
A learning task using GNNs can be viewed as extracting the features of the graph nodes, which contribute to further processing in the learning process.
The learning of graph nodes feature is denoted as:
\begin{equation}
\bm{X}^{(out)} = h(\bm{A},\bm{X}^{(in)})
\end{equation}
in which $\bm{A} \in \mathbb{R}^{N \times N}$ denotes a adjacent matrix of a graph, namely the structure of the graph, while $\bm{X}^{(out)} \in \mathbb{R}^{N \times d_{out}}$ and $\bm{X}^{(in)} \in \mathbb{R}^{N \times d_{in}}$ denotes the input and output feature matrix.
The operator of $h(\cdot,\cdot)$ is usually considered as a graph filter or convolution operator when using the nodes feature and structure information as input while the new node feature as output.

A typical GNN structure with $L$ graph filtering layers and $L-1$ activation layers, we denote the $i$-th graph filtering layer and activation layer as $h_i(\cdot)$ and $\alpha_i(\cdot)$, respectively.
Meanwhile, $\bm{X}^{(i)}$ denotes the output of the graph filtering layer, thus $\bm{X}^{(0)}$ means the original node feature matrix $\bm{X}$.
The output dimension of the $i$-th filtering layer is $d_i$.
Since the structure of the graph is constant is this work, then we have $\bm{X}^{(i)} \in \mathbb{R}^{n \times d_i}$.
Each node feature vector is embedded from the previous output of the activation layer, thus we have:
\begin{equation}
\bm{Z}^{(l)}=\bm{A}\bm{X}^{(l-1)}\bm{W}^{(l-1)}, \quad \bm{X}^{l} = \alpha_i(\bm{Z}^{l})
\end{equation}
in which $\bm{X}^{l} \in \mathbb{R}^{N \times d_l}$ denotes the embeddings in the $l$-th layer, $\bm{X}^{(0)} = \bm{X}$ denotes the input of the model when $l = 0$.
The learnable matrix of $\bm{W}^{(l)} \in \mathbb{R}^{d_l \times d_{l+1}}$ is used for feature transform in later layers. 


\subsubsection{Spectral Graph Theory}
The spectral graph theory incorporates traditional signal processing tools into analyzing graph structures.
The Fourier analysis used in graph is based on the graph Laplacian matrix:
\begin{equation}\label{equ:lap}
\bm{L} = \bm{D} - \bm{A}
\end{equation}
in which $\bm{A}$ denotes the adjacency matrix while $\bm{D}$ means the diagonal degree matrix of $A$, namely:
\begin{equation}\label{equ:lap_d}
\bm{D}_{ii} = \sum_j \bm{A}_{ij}
\end{equation}
The normalized version of graph Laplacian matrix is:
\begin{equation}\label{equ:norm_lap}
\hat{\bm{L}} = \bm{I} - \bm{D}^{-\frac{1}{2}}\bm{A}\bm{D}^{-\frac{1}{2}}
\end{equation}
which is decomposed as $\hat{\bm{L}} = \bm{U}\bm{\Lambda}\bm{U}^T$ since it is semidefinite matrix. $\bm{U}$ is the orthonormal eigenvector matrix while $\bm{\Lambda}$ denotes the diagonal matrix with $N$ eigenvalues of $\{\lambda_1, \ldots, \lambda_N\}$.

Consider the graph data of $\bm{X}$, the graph Fourier transform denoted as $\hat{\bm{X}} = \bm{U}^T\bm{X}$, which illustrate the Fourier transform coefficients.
Hence we have the graph convolution with input data $\bm{X}$ and filter $h$ as:
\begin{equation}\label{equ:graph_fourier}
\bm{X} \ast_\mathcal{G} \bm{h} = \bm{U}((\bm{U}^T\bm{h})\odot(\bm{U}^T\bm{X})) = \bm{U}\hat{\bm{h}}\bm{U}^T\bm{X}
\end{equation}
in which $\odot$ means the element-wise multiplication while $\hat{\bm{h}} = \mbox{diag}(\hat{h}_1, \ldots, \hat{h}_N)$ denotes the coefficients of the spectral filter.
With the filter coefficients, the inverse graph Fourier transform reconstructs the graph data of $\bm{X}$, which is denoted as $\bm{X}'$ as the filtered graph signal in a GNN framework.

In Equation \ref{equ:graph_fourier}, the convolution operator of $\bm{U}\hat{\bm{h}}\bm{U}^T$ has a high learning complexity depends on the node size of $N$. 
The forward propagation needs to perform matrix decomposition with a computing complexity of $\mathcal{O}(N^2)$.
Defferrard et al. \cite{defferrard2016cnngflsf} proposed a Chebyshev polynomials to approximate the filtering operation as follows:
\begin{equation}
\bm{U}\hat{\bm{h}}\bm{U}^T\bm{X} = \sum_{i=0}^K \theta_iT_i(\hat{\bm{L}}')\bm{X}
\end{equation}
in which 
\begin{equation}
\hat{\bm{L}}' = \frac{1}{\lambda_{max}}\hat{\bm{L}}-\bm{I}
\end{equation}
denotes the scaled normalized Laplacian (with eigenvalues ranging $[-1, 1]$), and $\lambda_{max}$ is the maximum eigenvalue, and $T_i(x)$is the Chebyshev polynomials which is recursively defined by
\begin{equation}\label{equ:cheb}
\left\{ \begin{array}{l}
T_0(x) = 1 \\ 
T_1(x) = x \\ 
T_i(x) = 2xT_{i-1}(x) - T_{i-2}(x)
\end{array}\right.
\end{equation}
In this work, the filtering layers we used are based on the Chebyshev polynomials function, termed as ChebNet Layers, present in the following sections.

\subsection{Transfer Learning}
\textit{Transfer learning} is a critical deep learning strategy, which has become popular in recent NLP and image processing applications. 
Transfer learning reuses the knowledge by applying the knowledge obtained from solving one problem to another different but related problem, transferring knowledge from the source domain to the target domain. 
The learning strategy will substantially impact a variety of areas when training data labels are insufficient. 
The core idea and learning process of transfer learning is that tasks \#1 and Task \#2 need to be learned, which are similar but not the same. The purpose of transfer learning is to discover and transfer the potential knowledge and connections in these two tasks to improve the learning performance in Task \#2 and make the training process in Task \#2 more efficient \cite{tan2018tf}.

There are four categories of deep transfer learning, including instance-based, relation-based, feature-based, and parameter-based transfer learning \cite{tan2018tf}.
This work focuses on the parameter-based deep transfer model with the GNN framework.
For datasets of HAR tasks, the sensor types, ambient environments, and discrepancies of different subjects' specified locations bring the data and dataset heterogeneity.
However, different datasets hold similar information to describe the human action state when describing similar activities.
These correlations of the different datasets in describing human activities make the transfer learning technique an appropriate tool in HAR, especially in the few-shot learning occasions.
This work uses GNN as our deep parameter-based transfer learning model to validate the transferability in sensor-based HAR as a typical mobile biomedical computing application.

\begin{figure*}[]
\centering
\includegraphics[width=7in]{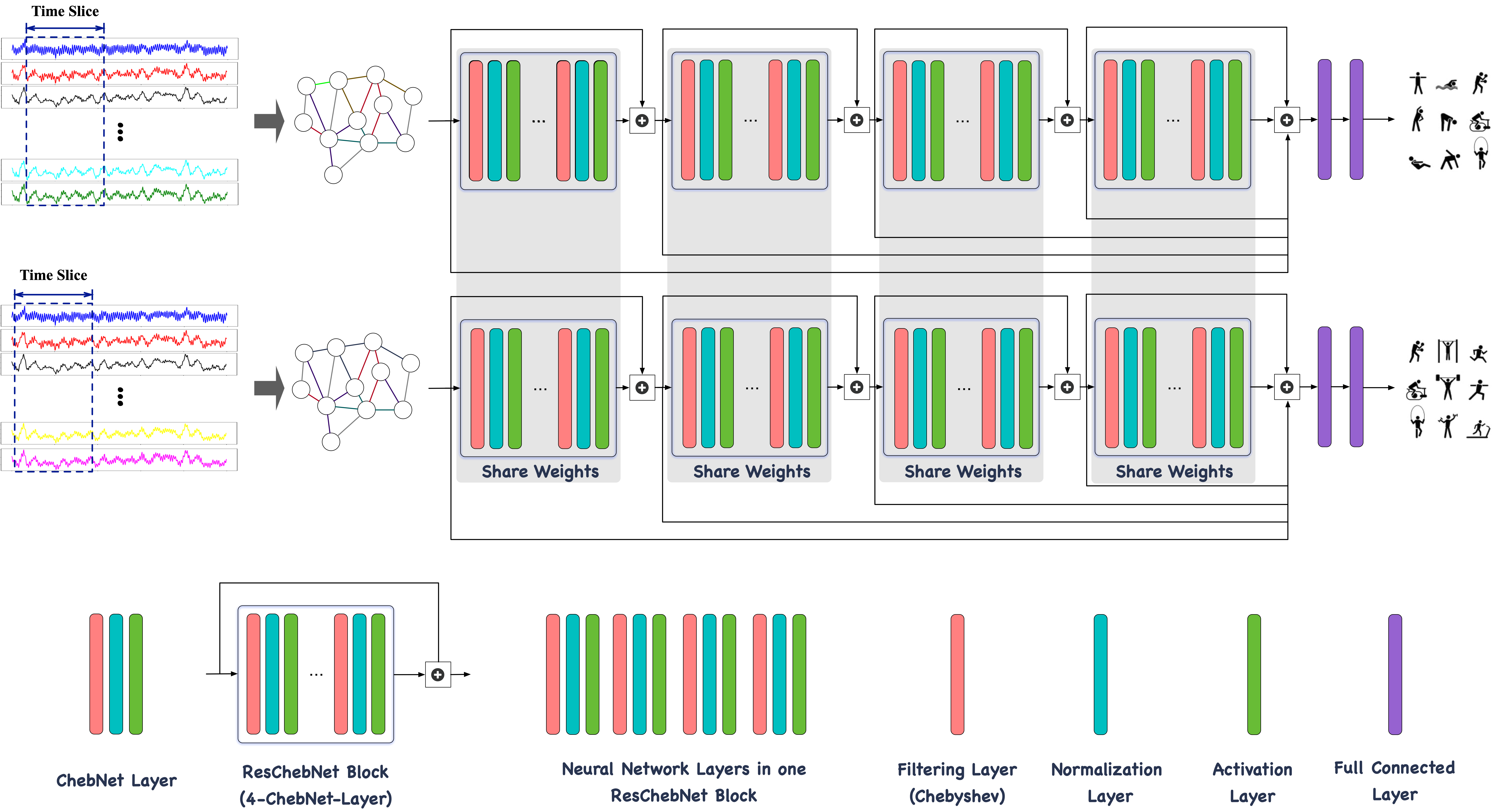}
\caption{The architecture of the ResGCNN and its transfer learning approach. The source and the target share the same set of model parameters. The four ResChebNet blocks with residual structure generate deep representations. Two fully connected layers are added to the four blocks in the source model. Each block contains four ChebNet layers composed of one Chebyshev graph filtering layer, one graph normalization layer, and one activation layer. Thus the parameters are trained with the source dataset in a supervised way. The trained parameters, namely the weights of the blocks, are transferred in the target model for feature extraction, two new fully connected layers are added in the target HAR task. The target dataset is used for training/testing to evaluate the transfer learning model with the ResGCNN structure.}
\label{fig_resgcn}
\end{figure*}

\section{Methodology}\label{sec:preliminaries}

This section presents a novel ResGCNN framework with transfer learning applications in sensor-based HAR tasks.
A parameter-based transfer strategy is proposed for activity recognition between different HAR datasets, which illustrates that the proposed ResGCNN-based deep transfer learning approach shows excellent transfer few-shot learning ability.
We first give the schematics of the ResGCNN framework, then give the detailed descriptions for the main component elements and the structure parameters, and finally give the training/testing algorithms for HAR.

\subsection{ResGCNN Framework}
In this work, we consider the ResGCNN framework in two cases.
Firstly, the ResGCNN framework is used for sensor-based activity recognition in one individual dataset, without transfer learning.
Secondly, we consider the deep transfer learning between different datasets with different sensor settings or activity types.
The HAR-ResGCNN deep transfer learning includes four main stages, including: 
\begin{enumerate}
\item Dataset preparation stage includes signal pre-processing, sliding window-based segmentation, and data preparation which converts multi-channel signals into graph data. This step should apply to all the involved datasets.
\item Model training with the source dataset, which optimizes the weights of each layer in the HAR-ResGCNN structure.
\item Parameter sharing uses the same residual graph network structure with the trained weights and ignores the last fully connected layers.
\item Model adaptation with the target dataset samples optimizes the full connected layers in the target model. 
\end{enumerate}
The HAR-ResGCNN deep transfer learning workflow schematic is illustrated in Figure \ref{fig_resgcn}.
Meanwhile, the framework's non-transfer only needs dataset preparation processing, and model training/testing as machine learning tasks do.


\subsection{Pre-processing and Data Preparation}
The multi-channel signals in the HAR dataset are firstly segmented into temporal slices.
Each slice contains multiple channels acquired from the wearable sensors. We treat the signals from a different sensor or location equally without discrimination, which are used to build the graph data. 
Consider the slice with multiple channels denoted as $\bm{X} = \{\bm{x}_1, \cdots, \bm{x}_M\}$, in which $M$ denotes the channel number of the signals.
The graph data include the graph signals $\{\bm{x}_1, \cdots, \bm{x}_M\}$ with the corresponding adjacency matrix $\bm{A}$ to form the connection relationship.
In this work, we build the adjacency matrix $\bm{A}$ with the Pearson's correlation coefficients as:
\begin{equation}\label{equ:pcc}
\rho(\bm{x}_i, \bm{x}_j) = \frac{E[(\bm{x}_i-\mu_{\bm{x}_i})(\bm{x}_j-\mu_{\bm{x}_j})]}{\sigma_{\bm{x}_i} \sigma_{\bm{x}_j}}
\end{equation}
in which $P$ denotes the length of the signal slice, while $m$ denotes the channel number from $M$ channels.
We build the HAR non-directed graph adjacency matrix with the Pearson's correlation coefficients and a threshold value $\psi$, thus the adjacency matrix of the graph data is decided by:
\begin{equation}\label{equ:adj}
\bm{A}_{i,j} = \left\{ \begin{array}{rcl}
0&\mbox{for}&\rho(\bm{x}_i, \bm{x}_j) < \psi \\
1&\mbox{for}&\rho(\bm{x}_i, \bm{x}_j) \geq \psi 
 \end{array}\right.
\end{equation}
The inputs of the proposed framework are turned into HAR graph data tuples as $\{\bm{X}, \bm{A}\}$.

\subsection{Network Structures}
We build the ResChebNet Blocks with four ChebNet layers (each containing one Chebyshev graph filtering layer with sequential graph normalization and activation layers) and the inter-block residual structure. 
The ResGCNN framework includes four ResChebNet blocks and two extra fully connected (FC) layers.
Simultaneously, an intra-block residual structure is involved, which adds the inputs of the four blocks to the output of the last block as the final output of the ResChebNet blocks.

\subsubsection{ResChebNet Block}

Let $\bm{W}$ denote the optimal adjacency matrix which formed the Laplacian matrix $\bm{L}^*$.
With the Laplacian matrix $\bm{L}^*$ we have the spatial graph filtering function as $h(\bm{L}^*)$, while its corresponding orthonormal eigenvector matrix of $\bm{U}^*$ and diagonal matrix of $\bm{\Lambda}^* = \mbox{diag}\{\lambda_0^*, \ldots, \lambda_N-1^*\}$.
Using the $K$-order Chebyshev polynomials as in Equation \ref{equ:cheb}, the graph filtering function can be approximated with
\begin{equation}\label{equ:output_cheb}
h(\bm{\Lambda}^*) = \sum_{k=0}^{K-1} \theta_k T_k(\tilde{\bm{\Lambda}}^*) = \sum_{k=0}^{K-1} \theta_k T_k(\frac{2\bm{\Lambda}^*}{\lambda^*_{max}} - \bm{I}_N)
\end{equation}
Thus, the output for a graph data input $\bm{X}$ is calculated with:
\begin{equation}
\begin{split}
\bm{y} = \bm{U}^* h(\bm{\Lambda}^*) \bm{U}^{*T}\bm{X} \\
=\sum_{k=0}^{K-1}\bm{U}^* \left[ \begin{array}{ccc}
   \theta_kT_k(\lambda_0^*) & \cdots & 0 \\
   \vdots & \ddots & \vdots \\
    0 & \cdots & \theta_kT_k(\lambda_{N-1}^*)  
   \end{array} \right]
\bm{U}^{*T}\bm{X} \\
= \sum_{k=0}^{K-1}\theta_kT_k(\tilde{\bm{L}}^*)\bm{X}
\end{split}
\end{equation}
in which $\tilde{\bm{L}}^*= 2\bm{L}^*/\lambda_{max}^* - \bm{I}_N$.

We term one graph filtering layer with its corresponding normalization layer and activation layer as a ChebNet layer, as shown in Figure \ref{fig_resgcn}.
The normalization layer used is based on the \textit{GraphNorm} layer proposed in \cite{pmlr-v139-cai21e}, which shows a fast converge speed compared to other GNN normalization strategies.
The activation function we choose is the Leaky Rectified Linear Unit (Leaky ReLU) function \cite{xu2015empirical}, with which an activation layer is added in each ChebNet layer.
We build one ResChebNet block with four sequential ChebNet layers with an inter-block residual structure.
The intra-network-block residual structure adds the block's inputs to the output of the last ChebNet layer.
As shown in Figure \ref{fig_resgcn}, the ResChebNet block's output equals the sum of block input and its last activation layer's output.
Consequently, we have a 12 layer residual structure in each block.

\subsubsection{Inter-Block Residual Structure}
The GCNN is essentially a Laplacian smoothing process, as the previous investigations show from a dynamic system viewpoint.
In the parameter tuning process of deep learning, the number of tunable parameters (weights and bias) increases when the number of layers is increasing, which causes the global and smooth degree of the nodes representations to increase \cite{chen2019over}.
The convolutional operations that make the nodes' representations approximate to each other is termed an effect of over-smoothing, which makes the dense part lose distinguish ability while the sparse part gain limited information.
To overcome the over-smoothing defects in deep GCNN, we were involved in a similar residual mechanism validated in\cite{chen2020revisiting, pei2020resgcn}, namely the inter-network-block residual structure as shown in Figure \ref{fig_resgcn}.
The inter-network-block residual structure uses the sum of the inputs of the four blocks and the output of the last block's activation layer output as the final extracted features.

\subsubsection{Classification}
In the last part of the proposed ResGCNN, we employ the softmax layer as the HAR classifier.
We add an FC layer to reduce the extracted features learned by the graph layers. 
The outputs of the FC layer are used as the features for classification using a softmax classifier layer:
\begin{equation}\label{equ:softmax}
\bm{P} = \mbox{softmax}(\bm{W}_{\mbox{softmax}}\bm{F}_{\mbox{FC}})
\end{equation}
where $\bm{P} = \{\bm{P}_1, \bm{P}_2, \ldots, \bm{P}_\mathcal{S}\}$ with $\bm{P}_s (s= 1, \ldots, \mathcal{S})$ represents the the predicted probability of the $s$-th activity types, while $\bm{W}_{\mbox{softmax}}$ denotes the \textit{softmax} layer parameters while $\bm{F}_{\mbox{FC}}$ denotes the features output of the FC layer.
We use the \textit{cross-entropy} error over all labeled samples as the loss function:
\begin{equation}\label{equ:loss}
\mathcal{L}_s = \mbox{cross-entropy}_s= -\sum_{c=1}^{\mathcal{C}} y_s log(P_s^{(c)}) 
\end{equation}
in which $P^{(c)}$ denotes the prediction probability of activity class of $c$.
A lower entropy error $\mathcal{L}_s$ indicates higher activity recognition accuracy.

\begin{table}[]
\caption{Data Information Used in This Work}
\begin{center}
\begin{tabular}{ccccc}
\toprule
\toprule
Block & Layer & Operator & Graph Filtering & Norm Size\\
 &  &  & Parameters Size & \\
 \midrule
1st & 1&ChebNet Layer&128$\times$256$\times$2&256 \\
1st & 2&ChebNet Layer&256$\times$512$\times$3&512 \\
1st & 3&ChebNet Layer&512$\times$256$\times$3&256 \\
1st & 4&ChebNet Layer&256$\times$128$\times$2&128 \\
\midrule
2nd & 5&ChebNet Layer&128$\times$256$\times$2&256 \\
2nd & 6&ChebNet Layer&256$\times$512$\times$3&512 \\
2nd & 7&ChebNet Layer&512$\times$256$\times$3&256 \\
2nd & 8&ChebNet Layer&256$\times$128$\times$2&128 \\
\midrule
3rd & 9&ChebNet Layer&128$\times$256$\times$2&256 \\
3rd & 10&ChebNet Layer&256$\times$512$\times$3&512 \\
3rd & 11&ChebNet Layer&512$\times$256$\times$3&256 \\
3rd & 12&ChebNet Layer&256$\times$128$\times$2&128 \\
\midrule
4th & 13&ChebNet Layer&128$\times$256$\times$2&256 \\
4th & 14&ChebNet Layer&256$\times$512$\times$3&512\\
4th & 15&ChebNet Layer&512$\times$256$\times$3&256 \\
4th & 16&ChebNet Layer&256$\times$128$\times$2&128 \\
\midrule
 - & 17&FC Layer&128$\times$64& - \\
 - & 18&Softmax Layer&64$\times$No. Labels& - \\
\bottomrule
\bottomrule
\label{tab_structure}
\end{tabular}
\end{center}
\end{table}

Mathematically, consider the overall loss denoted as $\bm{\mathcal{L}}$ with:
\begin{equation}
\bm{\mathcal{L}} = \mbox{cross-entropy}(\bm{1},\bm{1^P})
\end{equation}
where $\bm{1}$ and $\bm{1^P}$ denote the true activity (label) vector of training data and the predicted probability vector, respectively.
With the Back Propagation (BP) algorithm, the training with the annotated data samples dynamically learn the optimal adjacency matrix, the optimal FC layer parameters matrix, and the \textit{softmax} layer parameters denoted as $\bm{W}^*$, $\bm{W}_{\mbox{FC}}$, and $\bm{W}_{\mbox{softmax}}$, respectively. Using $\bm{W}$ as the optimal parameter tuple, the parameter updating process is denoted as:
\begin{equation}
\bm{W} = (1 - \rho) \bm{W} + \rho \frac{\partial \bm{\mathcal{L}}}{\partial \bm{W}}
\end{equation}
in which $\rho$ denotes the learning rate, while $\frac{\partial \bm{\mathcal{L}}}{\partial \bm{W}}$ is:
\begin{equation}
 \frac{\partial \bm{\mathcal{L}}}{\partial \bm{W}} = 
 \left(  
 \begin{array}{ccc} 
 \frac{\partial \mathcal{L}}{\partial W_{11}} & \ldots &\frac{\partial \mathcal{L}}{\partial W_{1N}} \\
 \vdots & \ddots &\vdots \\
 \frac{\partial \mathcal{L}}{\partial W_{N1}}  &\ldots &\frac{\partial \mathcal{L}}{\partial W_{NN}} 
 \end{array} 
 \right)
\end{equation}
where the element of $\frac{\partial \mathcal{L}}{\partial W_{ij}} $ is calculated by:
\begin{equation}
\frac{\partial \mathcal{L}}{\partial W_{ij}} = \frac{\partial \bm{\mathcal{L}}(\bm{1}, \bm{1^P})}{\partial \tilde{\bm{L}}} \cdot \frac{\partial \tilde{\bm{L}}}{\partial W_{ij}}
\end{equation}



\subsection{Transfer Learning of ResGCNN}
In this work, we use the pre-trained blocks in ResGCNN structure performed on the source domain as the feature extractor in the target domain.
Mathematically, we define the learning task performed on the source domain (dataset) as $\mathcal{D}_\mathcal{S} = {\{\bm{\mathcal{X}}_\mathcal{S}, \bm{\mathcal{Y}}_\mathcal{S}\}}$, while the target domain as $\mathcal{D}_\mathcal{T} = {\{\bm{\mathcal{X}}_\mathcal{T}, \bm{\mathcal{Y}}_\mathcal{T}\}}$.
We perform transfer learning of ResGCNN with the following strategy:
\begin{enumerate}
\item Perform the learning task on $\mathcal{D}_{\mbox{source}}$ with randomly initialized parameters, in which the optimized $\{\bm{\Theta}_{\mbox{blocks}}, \bm{\Theta}_{\mbox{FC}_\mathcal{S}}, \bm{\Theta}_{\mbox{softmax}_\mathcal{S}}\}$ are achieved using training algorithm in Algorithm \ref{alg:resgcnn}.
\item Build the ResGCNN structure in the target domain with the pre-trained parameters of $\bm{\Theta}_{\mbox{blocks}}$, and randomly initialized  FC layer and softmax layer, the node number of the softmax layer is decided by the number of the activity class in $\mathcal{D}_\mathcal{T}$;
\item Perform the learning task on $\mathcal{D_T}$ using the same ResGCNN structure with fixed initialized block parameters of $\bm{\Theta}_{\mbox{blocks}}$, while randomly initialized FC layer and softmax layer.
\item Perform the training process with the target ResGCNN structure to get the optimized parameters of $\{\bm{\Theta}_{\mbox{blocks}_\mathcal{T}}, \bm{\Theta}_{\mbox{FC}_\mathcal{T}}, \bm{\Theta}_{\mbox{softmax}_\mathcal{T}}\}$ with the training set;
\item Evaluate the target ResGCNN with the optimized parameter set.
\end{enumerate}

\begin{algorithm*}[]
\label{alg:resgcnn}
\SetAlgoLined
\textbf{Input:} Multichannel wearable sensor signals, the activity labels corresponding to the signal segments, the number of Chebyshev polynomial order $K$, the learning rate $r$;\\
\textbf{Output:} The optimized weight matrix $\bm{W}$ and model parameters of ResGCNN;  \\
1: Randomly initialize model parameters in the ResGCNN;\\
2: Initialize the adjacency matrix $\bm{A}$ based on \ref{equ:pcc} and \ref{equ:adj}\\
3: \textbf{repeat}\\
4:$\quad$ Computing the degree matrix $\bm{D}$\\
5:$\quad$ Computing the Laplacian matrix $\bm{L}$ \\
6:$\quad$ Computing the normalized version of Laplacian matrix $\tilde{\bm{L}}$ \\
7:$\quad$ Computing the Chebyshev polynomial elements $T_k(\tilde{\bm{L}}), k = 0, 1, \ldots, K-1$;\\
8:$\quad$ Computing the output of Chebyshev filtering layers with $\sum_{k=0}^{K-1}\theta_kT_k(\tilde{\bm{L}})x$ as illustrated in \ref{equ:output_cheb};\\
9:$\quad$ Normalizing the graph filtering output and perform the calculation of Relu activation functions; \\
10:$\quad$ Computing the results of the full connection layer; \\
11:$\quad$ Computing the cross-entropy based loss function;\\
12: $\quad$ Updating the parameter matrix with $\bm{W} = (1 - \rho) \bm{W} + \rho \frac{\partial \bm{\mathcal{L}}}{\partial \bm{W}}$ and other parameters of the ResGCNN; \\
13:\textbf{until} the iterations satisfy predefined certain training epochs or loss convergence condition
\caption{Network Parameter Training of ResGCNN Model}
\end{algorithm*}


\section{Experiments}\label{sec:experiments}
In this work, to validate the effectiveness of the proposed framework, we perform the experiments with three datasets, including two public open benchmark dataset and one self-own dataset, including:
\begin{enumerate}
\item HAR with ResGCNN framework performed with the three datasets;
\item HAR with deep transfer learning of ResGCNN framework, performed with different transfer settings.
\end{enumerate}
Below we firstly describe the datasets we use, and then we present the detailed implementations.

\subsection{Datasets}
In this work, we use three sensor-based HAR datasets to validate the learning framework of the transfer learning ability; each dataset has different sensor locations and numbers of activity labels.
The two public open datasets of PAMPA2 and mHealth datasets are widely adopted in sensor-based HAR model evaluation.
The third dataset is also built with motion sensors to track the physical activity of the human body. 
We only give brief introductions for the three involved datasets. Details of the datasets were presented in the provided reference.

\subsubsection{PAMAP2-HAR Dataset}
The PAMAP2 HAR dataset includes data acquired from $9$ participants of $24$ to $30$ years old. The participants wore IMUs on their dominant-side wrist, ankle, and chest.
Each person performed activities including \textit{lying down}, \textit{sitting}, \textit{standing}, \textit{walking}, \textit{running}, \textit{cycling}, \textit{Nordic walking}, \textit{ascending stairs}, \textit{descending stairs}, \textit{vacuum cleaning}, \textit{ironing clothes} and \textit{jumping rope}. Each IMU contains two 3D-acceleration sensor, a gyroscope-sensor, a magnetometer sensor, with sampling frequency of $100$Hz. 
More details of the dataset can be referred from \cite{arif2015physical, reiss2012creating}.
We ignore the \textit{vacuum cleaning} and \textit{ironing clothes} activities in this work.

\subsubsection{mHealth-HAR Dataset}
The mHealth-HAR dataset includes data from 10 participants in an out-of-lab environment.
Each subject wore wearable sensors attached to the chest, right wrist, and left ankle. 
Physical activities of \textit{standing still}, \textit{sitting}, \textit{lying}, \textit{walking}, \textit{climbing stairs}, \textit{waist bends forward}, \textit{frontal elevation of arms}, \textit{knees bending}, \textit{cycling}, \textit{jogging}, \textit{running}, and \textit{jumping front back} are involved in the experiment.
The sampling rate of recorded data is 50Hz. Further details on the experimental data collection can be found in \cite{banos2014mhealthdroid, banos2015design}

\subsubsection{TNDA-HAR Dataset}
We use the IMU sensors to capture the physical activity information for HAR and topological nonlinear dynamics analysis, which we denote as the TNDA dataset \cite{4epb-pg26-21}.
In this dataset, 50 subjects were recruited with ages ranging from 20 to 35, with 25 females and 25 males. 
We use five sensors located at the left ankle, left knee, back, right wrist, and right arm. Each IMU sensor includes a 3-D accelerometer, gyroscope, and magnetometer. 
The activity labels including \textit{standing still}, \textit{sitting}, \textit{lying down}, \textit{walking}, \textit{running}, \textit{walking up stairs}, \textit{walking down stairs} and \textit{cycling}, each with around 120 seconds. 
The sampling rate of the sensors is 50Hz in the TNDA HAR dataset.

\begin{table*}[]
\caption{Data Information Used in This Work}
\begin{center}
\begin{tabular}{cccccc}
\toprule
\toprule
 Dataset & No. of Activity Labels & Activity Contents  &  No. of Channels & Sensors & No. of Segments \\
 \midrule
PAMAP2 & 10 & \textit{standing still}, \textit{sitting}, \textit{lying}, \textit{walking}& 27 & 9-D right waist & 11784 \\
 & & \textit{running}, \textit{cycling}, \textit{Nordic walking} &  &  9-D left ankle, &    \\
& & \textit{ascending stairs}, \textit{descending stairs}, \textit{roping}    &  & and 9-D back&    \\
\midrule
 mHealth & 12 & \textit{standing still}, \textit{sitting}, \textit{lying}, \textit{walking} & 21 & 9-D right waist,&   5361  \\
  &&   \textit{climbing stairs}, \textit{waist bends forward} &  &  9-D left ankle, &    \\
  &&  \textit{frontal elevation of arms}, \textit{knees bending} &  &  and 3-D chest &   \\ 
 & & \textit{cycling}, \textit{jogging}, \textit{running}, \textit{jumping front back} &  &  &   \\ 
\midrule
 TNDA-HAR & 8 & \textit{standing still}, \textit{sitting}, \textit{lying down} & 27  & 9-D right waist, & 29112 \\
   &  &  \textit{walking}, \textit{running}, \textit{walking up stairs} &  & 9-D left ankle,  &  \\
    &  & \textit{walking down stairs} and \textit{cycling} &  & and 9-D back &  \\
\bottomrule
\bottomrule
\end{tabular}
\end{center}
\label{table_dataset}
\end{table*}

\subsubsection{Pre-processing and Data Preparation}
To perform our experiments and corresponding comparisons, we adjust the datasets with the following processing. 
Firstly, we use 50Hz as our standard frequency rate. Thus, signals from the PAMAP2 dataset are downsampled from 100Hz to 50Hz, to keep the consistency of the sampling rate of signals.
Secondly, we use only the information of 3 sensors in the TNDA, PAMAP2 dataset, including the right waist, left ankle, and back, to keep the consistency of sensor locations. The sensing information of the body in mHealth is captured with a chest sensor, while the other two are from the back sensors.
Thirdly, we use a temporal size of 128 with 50\% is used to extract signal segments. Each segment includes 21, 27, and 27-dimensional signals from the mHealth, PAMAP2, and TNDA datasets, respectively.
Details of the dataset information used in the experiments are illustrated in Table \ref{table_dataset}.

\subsection{Implementations}
\subsubsection{Model Parameters}
In this work, we use the \textit{Pytorch} framework with the \textit{geometric} package to build the neural network model. 
For each evaluation, we perform 5-fold cross-validation and 80\% of the samples as training data while 20\% as the testing data.
In the experiments, the learning rate of the model is set as 0.001, and the batch size is set as 64.
We use a maximum of 120 training rounds and the \textit{Adam} stochastic optimization algorithm to optimize the network parameters. 
For each signal segment, the multi-channel signals are firstly converted into graph-based data as described in Equation \ref{equ:pcc} and \ref{equ:adj}, the correlation threshold is set as 0.2.
The implementation codes for the model and main experiments are accessible in \href{https://github.com/liaotian1005/gnn}{https://github.com/liaotian1005/gnn}.

\subsubsection{Evaluation Criterial}
For the classification task, we use classification accuracies, recalls, F1-scores, and confusion matrix to illustrate the accomplished results.
For each activity class in the datasets, the predictions of the model were compared to the ground truth labels to calculate the numbers of true-positives (TP), true-negatives (TN), false-positives (FP), and false-negatives (FN). 
The overall accuracy equals 
\begin{equation}
\mbox{Accuracy} = \frac{\mbox{TN} + \mbox{TP}}{\mbox{TN} + \mbox{TP} + \mbox{FN} + \mbox{FP}}
\end{equation}
and the precision and recall of one typical class can be calculated by 
\begin{equation}
\mbox{Precision} = \frac{\mbox{TP}}{\mbox{TP} + \mbox{FP}} \quad
\mbox{Recall} = \frac{\mbox{TP}}{\mbox{TP} + \mbox{FP}}
\end{equation}
The F1-score is a balanced metic combination of both the precision and recall as:
\begin{equation}
\mbox{F1-score}=\frac{2*\mbox{Precision}*\mbox{Recall}}{\mbox{Precision} + \mbox{Recall}} * 100\%.
\end{equation}
The average values of the activity labels are used for the assessment of each experiment.
In addition, the confusion matrices are also involved in visualizing the performance of models.



\begin{figure*}[]
\centering
\includegraphics[width=7.2in]{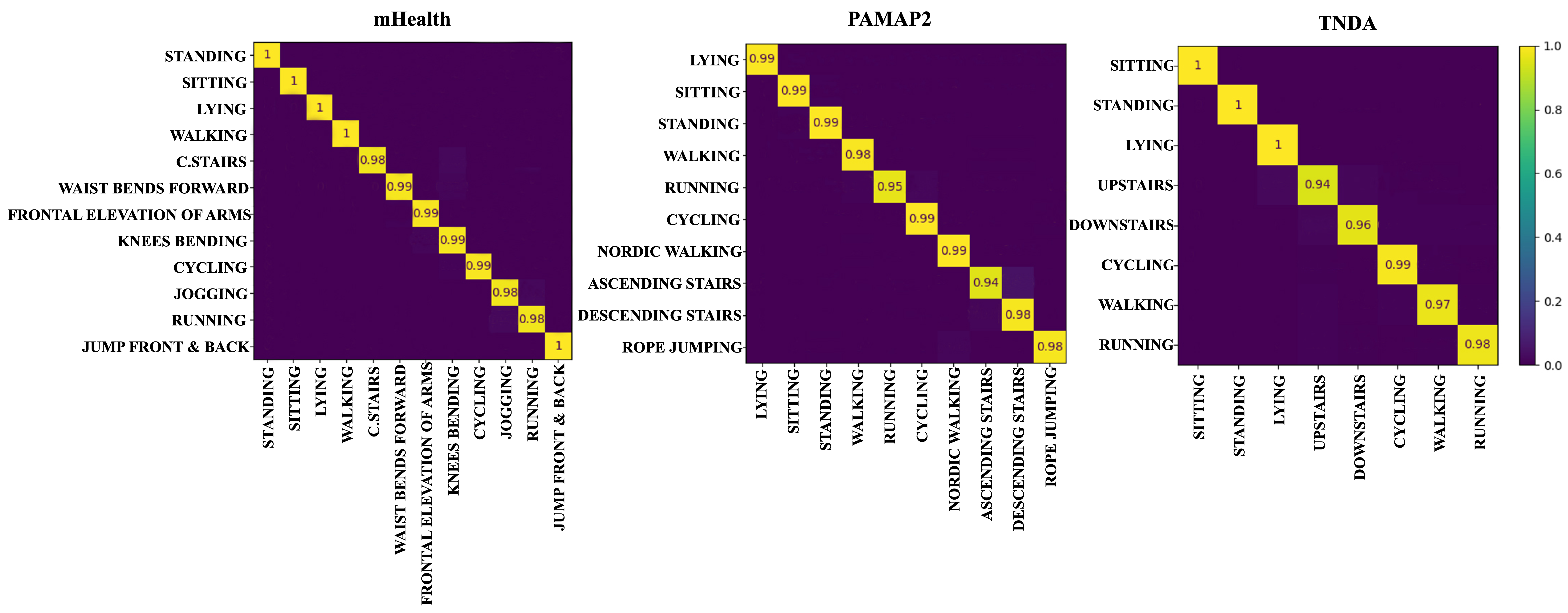}
\caption{The Confusion Matrices with ResGCNN Experiments with PAMAP2 (left), mHealth (middle), and TNDA (right).}
\label{fig_har_results}
\end{figure*}

\section{Results \& Discussion}\label{sec:experiments}
\subsection{HAR with ResGCNN}
\subsubsection{Achieved Results}
The experimental results achieved on the datasets of PAMAP2, mHealth, TNDA are illustrated in Table \ref{table_har_results}. 
The overall accuracies are 98.18\%, 99.07\%, and 97.97\% for the PAMAP2, mHealth, and TNDA, respectively.
Besides, the average precision, recall, and F1-score of the ten involved activities in PAMAP2 are 97.86\%, 97.89\%, and 97.86\%, respectively; the average precision, recall, and F1-score of the twelve involved activities in mHealth are 96.95\%, 99.09\%, and 99.13\%, respectively; while in our TNDA dataset, we achieve an average precision, recall and F1-score of 99.10\%, 97.89\%, and 97.89\% for the involved eight activities, respectively.
The confusion matrices for the three datasets for HAR are shown in Figure \ref{fig_har_results}.

\subsubsection{Comparison with Related Works}
There are a variety of sensor-based HAR frameworks proposed in previous pieces of literature, especially with the recent rapidly developed deep learning techniques. We present some of the typical works using similar experimental settings upon the two open datasets, namely PAMAP2 and mHealth, separately.

For the PAMAP2 dataset, Alejandro et al. \cite{baldominos2017feature} proposed a HAR genetic algorithm handcraft feature set (280 features) analysis framework, achieving an average accuracy of $97.45\%$; Alok et al. \cite{chowdhury2017physical} utilize weighted fusion with correlation-based feature selection upon extracted 135 features, the achieved mean F1-score is 92.32\%;
Di Wang et al. \cite{wang2017robust} proposed a robust activity recognition model by extracting 44 features per sensor. The result average accuracy is 94.76\%; Abdullah et al. \cite{awal2019optimization} incorporate a Bayesian optimization framework for feature selection in HAR with a 95.4\%. 
Despite the feature-learning approaches, the recent development of deep learning techniques brought a leap in the HAR performance. Typical works evaluated upon the PAMAP2 include: Shaohua et al. \cite{wan2021dlm} adopted CNN framework for HAR achieved an average accuracy of 91.00\%; Li et al. \cite{li2021deepfeature} validate a CNN-LSTM framework with the dataset by achieving an average accuracy ranging from 96.97\% to 97.37\%.

For the mHealth dataset, Alok et al. \cite{chowdhury2017physical} achieved a mean F1-score of 88.59\% based on feature-learning.
Vijay et al. \cite{semwal2021} proposed a hybrid deep learning model using ensemble learning approach achieving an average accuracy of 94.00\%, which involved comparison with their previous work using CNN \cite{semwal2017robust}, LSTM\cite{semwal2017robust}, Gated Recurrent Unit (GRU) \cite{semwal2017robust}, CNN–LSTM\cite{semwal2017robust}, CNN–GRU\cite{semwal2017robust} and GRU–CNN\cite{semwal2017robust}, with corresponding average recognition accuracies of 91.66\%, 86.89\%, 81.77\%, 91.66\%, 91.66\%, and 92.53\%, respectively. 
Lingjuan et al. \cite{10.1145/3132847.3132990} adopted a two-stage randomization-based scheme in LSTM-CNN structure to achieve a privacy-preserving collaborative deep learning model with an average accuracy of 92\%.
Sojeong et al. \cite{7727224} used both partial weight sharing and full weight sharing mechanisms with 2D-CNNs for HAR, achieved an average accuracy of 91.94\%.
The best-achieved results performed with mHealth is 99.20\% proposed in \cite{qin2020imaging}.

In summary, our HAR-ResGCNN approach achieves comparable results compared to the state-of-art frameworks in sensor-based HAR tasks.
However, in real-world wearable computing applications, the acquired sensor data might be lacking annotated information, as well as inconsistency in on-body wearing position or sensor types.
Thus, the transfer learning paradigm brought significant ease in dealing with such scenarios as experiments in the following sections.

\begin{table*}[]
\caption{Classification Results with the proposed HAR-ResGCNN Approach}
\begin{center}
\begin{tabular}{cccccccccc}
\toprule
\toprule
 Dataset & & PAMAP2  & & & mHealth  & &  & TNDA &  \\
 \midrule
Labels/Metrics & Pre(\%) & Recall(\%) & F1-Score(\%) &  Pre(\%) & Recall(\%) & F1-Score(\%) & Pre(\%) & Recall(\%) & F1-Score(\%) \\
\midrule
\textit{sitting} & 98.80 &98.97 &98.88&100 & 100 & 100 &100&100&100\\
\textit{standing} & 98.26&98.95&98.60 & 100 & 100 & 100 & 100 & 99.87&99.93\\
\textit{lying} &99.84 & 98.87& 99.35& 100 & 100 & 100 & 98.09& 100&99.03\\
\textit{upstairs} & 96.69 & 94.34&95.50 & - & - & - &95.33&94.15&94.74\\
\textit{downstairs}& 92.66&97.91&95.21 & - & - & - &96.59&95.81&96.19\\
\textit{cycling} &98.61&99.00&98.80&100& 99.00&99.50&97.17&98.77&97.96\\
\textit{walking} & 98.75 &98.20 &98.48& 100 & 100 & 100 & 97.61&96.88&97.25\\
\textit{running} & 98.98 & 95.41 & 97.16& 97.87 &   97.87 &  97.87  &98.36&97.62&97.99\\
\textit{Nordic walking} &97.75 &99.28&99.01& - & - & - & - & - & -\\
\textit{rope jumping} & 97.32 & 97.97 &97.64& - & - & - & - & - & - \\
\textit{C. stairs} &  - & - & -&  100 & 97.98 & 98.98 & - & - & -\\
\textit{waist bends forward} & - & - & -& 98.80 & 98.80 &  98.80 & - & - & -\\
\textit{frontal elevation of arms} &  - & - & -&99.06& 99.06 & 99.06 & - & - & -\\
\textit{knees bending} &  - & - & -&95.29& 98.78& 97.01 & - & - & -\\
\textit{jogging} &  - & - & -& 98.04 & 98.04 & 98.04 & - & - & - \\
\textit{jumping front \& back} &  - & - & -& 100 & 100 & 100 & - & - & - \\
\midrule
Average & 97.86 & 97.89& 97.86& 96.95 & 99.09& 99.13 & 99.10 & 97.89 & 97.89 \\
\midrule
Overall Accuracy &   & 98.18 & &  & 99.07 &  &  & 97.97 &  \\
\bottomrule
\bottomrule
\end{tabular}
\end{center}
\label{table_har_results}
\end{table*}

\subsection{Deep Transfer Learning of ResGCNN}
\subsubsection{Transfer Learning Conduces to Promotion in Model Training Efficiency}
\begin{figure}[]
\centering
\includegraphics[width=3.6in]{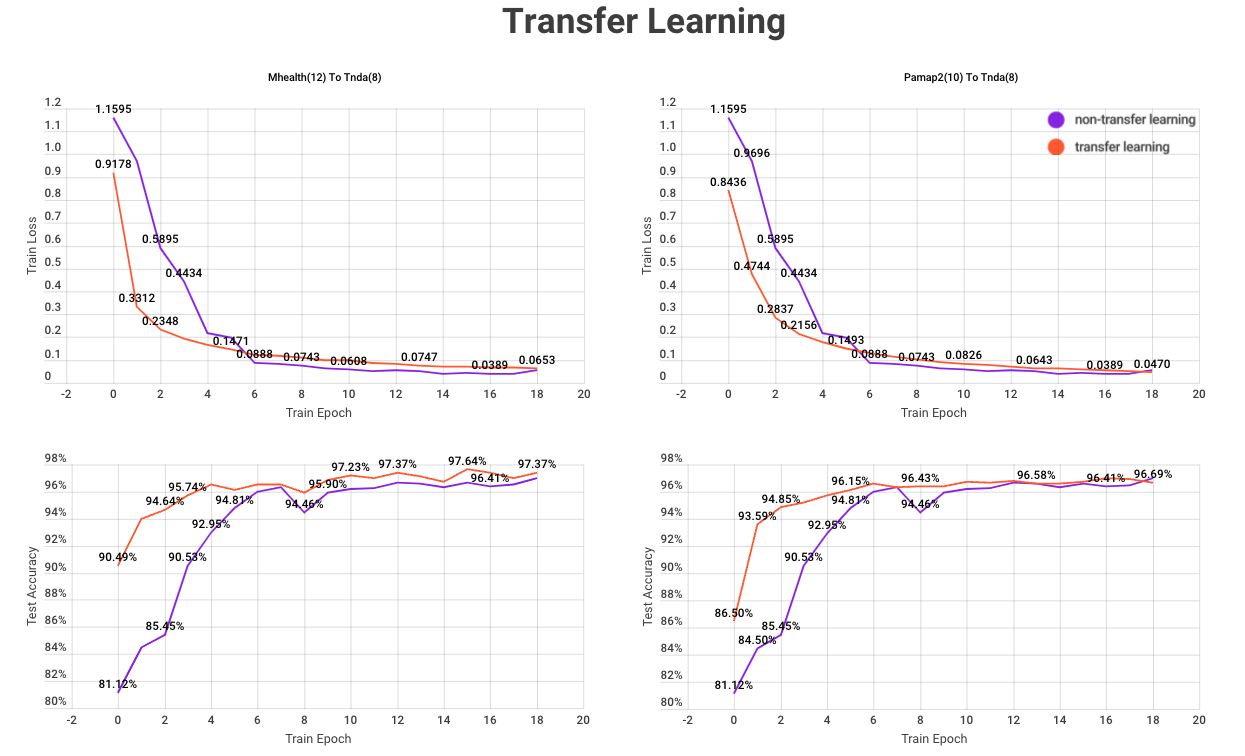}
\caption{Transfer learning performed upon TNDA dataset.}
\label{fig_tnda_curve}
\end{figure}

\begin{figure}[]
\centering
\includegraphics[width=3.6in]{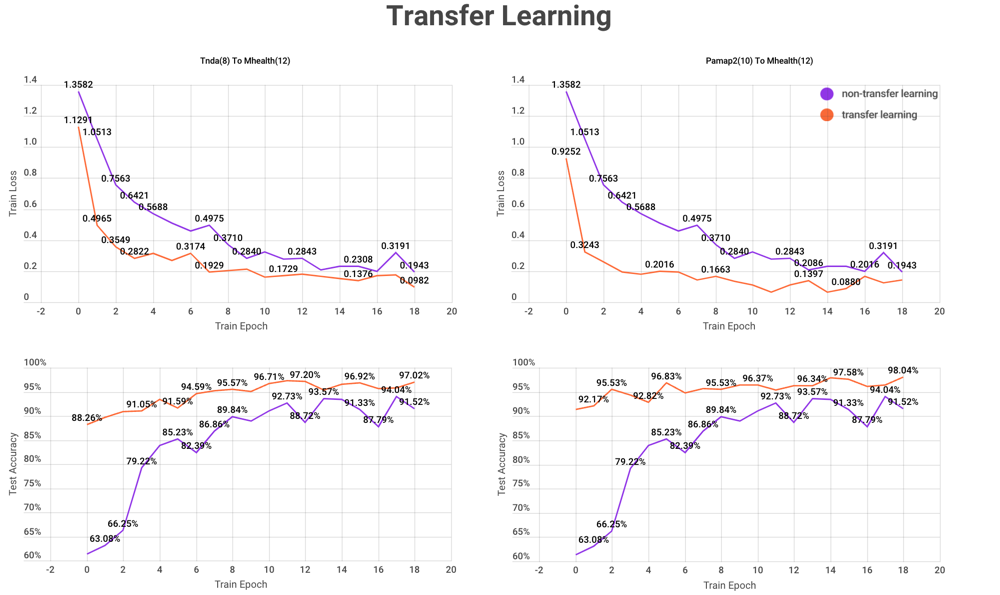}
\caption{Transfer learning performed upon mHealth dataset.}
\label{fig_mhealth_curve}
\end{figure}

In this work, we consider parameter-based transfer learning with the ResGCNN framework.
In this section, we perform the experiments upon the target dataset of TNDA and mHealth, which use an 80\% portion of data samples as the training set.
The initial parameters of the GNN layers in target ResGCNN in transfer learning experiments are based on the optimized parameters from the source ResGCNN.
The other layers of target ResGCNN are randomly initialized.
To keep consistency with the mHealth dataset (21 signal channels), the transfer learning tasks with mHealth use only the corresponding 21 channels as illustrated in Table \ref{table_dataset}.
The learning loss curves and test accuracy curves based on the training epochs are shown in Figure \ref{fig_tnda_curve} and \ref{fig_mhealth_curve}.

For the experiments performed upon TNDA as the target dataset, the transfer learning using mHealth and PAMAP2 as the source datasets separately.
We use M-to-T and P-to-T to denote the experiments using mHealth and PAMAP2 as the source dataset, respectively.
As illustrated in Figure \ref{fig_tnda_curve}, the test accuracies in TNDA dataset with transfer learning are 90.49\% and 86.50\% in the zeroth epoch case in M-to-T and P-to-T, respectively, which are better than the non-transfer case with 81.12\%. 
In M-to-T tasks, the average recognition accuracies achieve more than 96\% with four epochs in transfer, while with ten epochs in non-transfer cases.
In P-to-T tasks, the average recognition accuracies achieve more than 96\% with four epochs in transfer, while with nine epochs in non-transfer cases.

The transfer learning experiments upon mHealth performed as the target dataset uses TNDA and PAMAP2 as the source datasets, separately.
We use T-to-M and P-to-M to denote the experiments using TNDA and PAMAP2 as the source dataset, respectively.
As illustrated in Figure \ref{fig_mhealth_curve}, the test accuracies in TNDA dataset with transfer learning are 88.26\% and 92.17\% in the zero epoch case in T-to-M and P-to-M, respectively, which are better than the non-transfer case with 63.08\%. 
In the T-to-M task, the average recognition accuracy achieves 95\% with only about six epochs, while in the P-to-M task with only five epochs.
On the contrary, in the non-transfer case, the accuracy is still lower than 95\% within 20 epochs.

The results illustrate that the ResGCNN framework initialized with the transferred parameters shows a better learning ability and convergence speed than randomly initialized parameters.
Especially in the experiments using mHealth as the target dataset, the improvements are much more significant than that of the TNDA as target cases.
The reason is that in TNDA and mHealth, the complex degrees of activities are different: TNDA only contains eight activities while mHealth twelve activities, mHealth contains several much more subtle activities than in TNDA, such as \textit{frontal elevation of arms} and \textit{knees bending}.



\subsubsection{Transfer Learning Enhances Few-Shot Learning Ability}
The HAR application scenarios might suffer the lacking of annotated data samples, which brings the challenges of few-shot learning ability for the models.
In this section, we validate the performance of the proposed ResGCNN model and its transfer implementation in a few-shot learning case.
We consider two experiments with a training data sample ratio of 2.5\% and 5\% performed on the target dataset of PAMAP2, mHealth, and TNDA, while the other two as the source dataset.
The comparisons of the results in transfer \& non-transfer learning in few-shot cases are illustrated in Table \ref{tab_few_shot}.

\begin{table}[]
\caption{Transfer Learning Experiments and Comparisons in Few-Shot Learning}
\begin{center}
\begin{tabular}{cccccc}
\toprule
\toprule
TF Settings  & Non-TF(5\%) & TF(5\%)  & Non-TF (2.5\%)   & TF (2.5\%) \\
\midrule	
T-to-P &82.45\%&87.09\%& 81.32\%& 82.89\% \\
M-to-P&82.45\% &87.67\%& 81.32\%&85.20\% \\%
\midrule
T-to-M&57.89\% &86.41\% & 35.42\%&54.67\% \\
P-to-M &57.89\%&88.13\%& 35.42\%&67.29\%  \\
\midrule
M-to-T &89.89\% &93.37\%  & 86.39\% & 90.05\% \\
P-to-T &89.89\%& 92.67\% &  86.39\% & 92.84\%\\
\bottomrule
\bottomrule
\end{tabular}
\end{center}
\label{tab_few_shot}
\end{table}

\begin{figure*}[]
\centering
\includegraphics[width=7in]{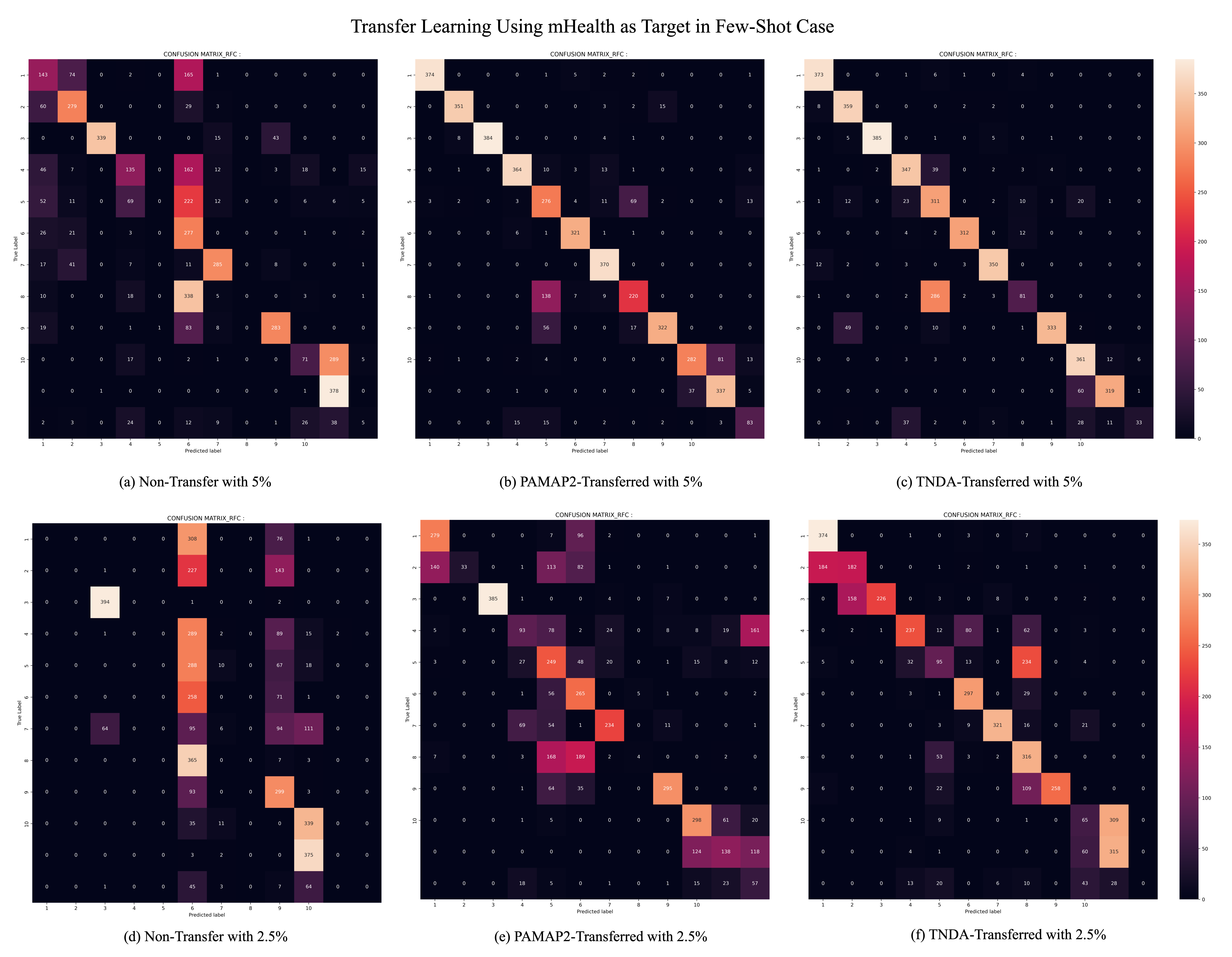}
\caption{The Confusion Matrices for Few-Shot Experiments Performed on the Target Dataset of mHealth.}
\label{fig_trans_mHealth}
\end{figure*}

In the 5\% experiments, the supervised training phase use 5\% of the object dataset while 80\% for testing.  
When using PAMAP2 as the target dataset, the non-transferred target ResGCNN achieves an overall accuracy of 82.45\%, while the transfer learning tasks of T-to-P (TNDA transferred to PAMAP2) and M-to-P (mHealth transferred to PAMAP2) reach overall accuracies of 87.09\% and 87.67\%, respectively. The transfer learning increases average accuracies by 4.64\% and 5.22\% with TNDA and mHealth, respectively.
When using mHealth as the target dataset, the non-transferred target ResGCNN achieves an overall accuracy of 57.89\%, while the transfer learning tasks of T-to-M and P-to-M accomplish overall accuracies of 86.41\% and 88.13\%, respectively. The transfer learning increases average accuracies by 28.52\% and 30.24\% with TNDA and PAMAP2, respectively.
When using TNDA as the target dataset, the non-transferred target ResGCNN achieves an overall accuracy of 89.89\%, while the transfer learning tasks of M-to-T and P-to-T achieve overall accuracies 93.37\% and 92.67\%, respectively. The transfer learning increases average accuracies by 3.48\% and 2.78\% with mHealth and PAMAP2, respectively.

Furthermore, we consider a more extreme case with reducing the training size from 5\% to 2.5\%.
In the 2.5\% experiments, the supervised training phase use 2.5\% of the object dataset while 80\% for testing.  
When using PAMAP2 as the target dataset, the non-transferred target ResGCNN achieves an overall accuracy of 81.32\%, the transfer learning tasks of T-to-P and M-to-P accomplish overall accuracies of 82.89\% and 85.20\%, respectively. The transfer learning increases average accuracies by  1.57\% and 3.88\% with TNDA and mHealth, respectively.
When using mHealth as the target dataset, the non-transferred ResGCNN overall accuracy is 35.42\%, the transfer learning tasks of T-to-M and P-to-M accomplish overall accuracies of 54.67\% and 67.29\% respectively. The transfer learning increases average accuracies by 19.25\% and 31.87\% with TNDA and PAMAP2, respectively.
When using TNDA as the target dataset, the non-transferred ResGCNN achieves an overall accuracy of 86.39\%, while the transfer learning tasks of M-to-T and P-to-T, we achieve the overall accuracies of 90.05\% and 92.84\%, respectively. The transfer learning increases average accuracies by 3.66\% and 6.45\% with mHealth and PAMAP2, respectively.

\subsubsection{Meta-Learning Ability of ResGCNN}
With the ResGCNN framework, we perform the transfer learning tasks in two few-shot learning cases, as presented in the above section.
As the illustrated confusion matrices in Figure \ref{fig_trans_mHealth}, we can see that the proposed ResGCNN model shows a certain degree of meta-learning ability, which improves the model learning ability in 'unseen' activities, especially in the few-shot learning case with limited annotated samples.
We use the confusion matrices performed upon the mHealth dataset as illustrated in Figure \ref{fig_trans_mHealth} as the example since it has a higher activity complex degree.

Consider the activity recognition tasks of \textit{c.stairs} (5th diagonal element), \textit{knees bending} (8th diagonal element), \textit{jump front} \& \textit{back} (12th diagonal element) in the 5\%-case of mHealth, with the P-to-M and T-to-M transfer learning, there exists a significant increase in the recognition ability of the activities.
The similar increases happen on 8 activity types in the 2.5\%-case of mHealth, including \textit{standing, sitting, walking, c.stairs, frontal elevation of arms, knees bending, running}, and \textit{jump front} \& \textit{back}.
Moreover, for the TNDA-transferred experiments, even though the TDNA dataset contains a limited number of activities, the ResGCNN framework still improves the recognition accuracy for the activities not included activities.
The phenomenons illustrated show our HAR-ResGCNN approach is capable of meta-learning ability in HAR tasks, which might help in understanding unseen physical activities in real-world HAR systems and applications.

\section{Conclusion}\label{sec:conclusion}
The sensor-based HAR is a representative mobile computing scenario in biomedical information acquisition and analysis.
Current models used in HAR might be impacted by the modalities variation when different types of sensors are involved and short of annotated data samples.
These characters restrict the recognition accuracy and intelligence level of the mobile computing systems involved in wearable motion sensor devices.
To handle the modality variation and annotated sample deficiency problems, we propose a deep transfer learning model for sensor-based HAR tasks, namely HAR-ResGCNN, in this work.
The ResGCNN structure is a multi-layer neural network composed of GNN with Chebyshev filtering functions and residual structures, which is designed to learn sensor signal representations and recognize human activities.
The significances of this work are twofold: 1) the ResGCNN framework shows excellent classification ability in HAR tasks; 2) the ResGCNN framework is used to establish a deep transfer learning analytical method for inter-dataset application HAR tasks.
Experiments performed on two open benchmark datasets and one self-acquired dataset shows that the ResGCNN framework has comparable recognition ability compared to the state-of-art models, while the transfer learning with ResGCNN shows great few-shot learning ability in distinguishing activity classes.
Our proposed approach is a decent solution to tackle sensor modalities variation and annotated data deficiency problems and is supposed to be a promising choice for sensor-based HAR tasks and mobile learning applications.

\bibliographystyle{IEEEtran}
\bibliography{gnn4siathar}


%





\ifCLASSOPTIONcaptionsoff
  \newpage
\fi

\end{document}